\documentclass[conference,a4paper]{IEEEtran}

\IEEEoverridecommandlockouts

\usepackage{cite}
\usepackage{amsmath,amssymb,amsfonts}
\usepackage{algorithmic}
\usepackage{graphicx}
\usepackage{textcomp}
\usepackage{xcolor}
\usepackage{float}
\usepackage{tikz}
\def\BibTeX{{\rm B\kern-.05em{\sc i\kern-.025em b}\kern-.08em
    T\kern-.1667em\lower.7ex\hbox{E}\kern-.125emX}}

\newcommand\copyrighttext{%
  \footnotesize \textcopyright “Copyright 2019 IEEE. Published in the Digital Image Computing: Techniques and Applications, 2020 (DICTA 2020), 30 November – 2 December 2020 in Melbourne, Australia. Personal use of this material is permitted. However, permission to reprint/republish this material for advertising or promotional purposes or for creating new collective works for resale or redistribution to servers or lists, or to reuse any copyrighted component of this work in other works, must be obtained from the IEEE. Contact: Manager, Copyrights and Permissions / IEEE Service Center / 445 Hoes Lane / P.O. Box 1331 / Piscataway, NJ 08855-1331, USA. Telephone: + Intl. 908-562-3966.”}
\newcommand\copyrightnotice{%
\begin{tikzpicture}[remember picture,overlay]
\node[anchor=south,yshift=10pt] at (current page.south) {\fbox{\parbox{\dimexpr\textwidth-\fboxsep-\fboxrule\relax}{\copyrighttext}}};
\end{tikzpicture}%
}

\begin{document}

\title{One-Shot learning based classification for segregation of plastic waste\\
}

\author{\IEEEauthorblockN{Shivaank Agarwal}
\IEEEauthorblockA{\textit{Department of Computer Science} \\
\textit{BITS Pilani} \\
Hyderabad, India. \\
shivaank.agarwal@gmail.com}
\and
\IEEEauthorblockN{Ravindra Gudi}
\IEEEauthorblockA{\textit{Department of Chemical Engineering} \\
\textit{IIT Bombay}\\
Mumbai, India. \\
ravigudi@iitb.ac.in}
\and
\IEEEauthorblockN{Paresh Saxena}
\IEEEauthorblockA{\textit{Department of Computer Science} \\
\textit{BITS Pilani} \\ 
Hyderabad, India. \\
psaxena@hyderabad.bits-pilani.ac.in}
}

\maketitle
\copyrightnotice

\vspace{-4mm}
\begin{abstract}
The problem of segregating recyclable waste is fairly daunting for many countries. This article presents an approach for image based classification of plastic waste using one-shot learning techniques. The proposed approach exploits \mbox{discriminative} features generated via the siamese and triplet loss convolutional neural networks to help differentiate between 5 types of plastic waste based on their resin codes. The approach achieves an accuracy of 99.74\% on the WaDaBa Database \cite{WaDaBa1}. 
\end{abstract}


\section{Introduction}
As stated by the Society of the Plastics Industry, Inc. (SPI), the purpose of plastic resin code is to provide a consistent system to facilitate the recycling of post-consumer plastic waste \cite{plastic}. We propose the automatic classification of plastic waste based on its resin code using deep learning approaches.
\newline Deep Learning using convolutional neural networks has achieved significant breakthrough in tasks such as image classification and object detection \cite{yolo,ssd}. However, these algorithms do not perform well when less amount of data is available for training. In such circumstances, it is better to learn discriminative features to distinguish between classes. We propose using one-shot learning to generalise over the entire dataset, where the amount of data for certain classes is not sufficient to train traditional convolutional neural networks.
\newline One-shot learning methods directly learn the similarities between objects of the same class and also the differences between objects of different classes. In this article, we propose the use of siamese \cite{Koch} and triplet loss \cite{facenet} networks on the plastic waste images available in the WaDaBa database.
\begin{figure}[H]
\centerline{\includegraphics[width=55mm]{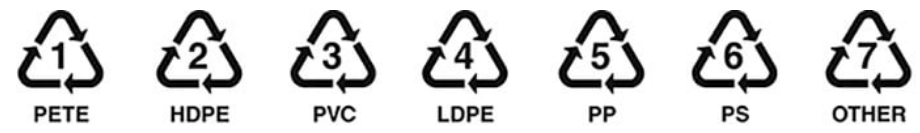}}
\caption{7 categories of Plastic Resin codes}
\label{resin_codes}
\end{figure}

\section{Related Work}
WaDaBa is the only dataset available which classifies plastic waste based on its resin code. There have been two previous works on this database. In \cite{WaDaBa1}, the classification was attempted using histogram analysis, and an accuracy of 75.68\% was achieved. In \cite{WaDaBa2}, the authors used image augmentation (rotation) to increase the size of the database from 4000 to around 140,000 images and then used CNNs for classification. They achieved an accuracy of 99.92\%. The authors classified images in 4 out of the 5 available categories in the database, due to a shortage of images in the category of other \mbox{plastics (category 7 in Fig. \ref{resin_codes}). }

While the above method already achieves high accuracy,  it requires substantial rotation augmentation. This does not add any unseen instances in the test data since deep CNNs are powerful enough to learn rotational invariant filters on their own, given a  sufficient number of training images. Hence the method may not work well on new instances of data that do not have a rotational variant in the training set. In this paper, we propose a model that does not require any augmentation to increase the size of the database, but achieves a comparable accuracy.

\section{WaDaBa Database}
The WaDaBa database contains 4000 images belonging to the 5 categories - PET (01-polyethylene terephthalate), PE-HD (02-high-density polyethylene), PP (05-polypropylene), PS (06-polystyrene) and other (07). Table \ref{table:images categ} shows the number of images per category in the database. Each image consists of a single object which is subjected to various degrees of deformations to simulate natural conditions. 
\begin{table}[htbp]
\caption{Images per category in WaDaBa}
\begin{center}
\begin{tabular}{@{} l c @{}}
\hline
\textbf{Category} & \textbf{No. of Images} \\ \hline
01 PET & 2200 \\ 
02 PE-HD & 600 \\ 
05 PP & 640 \\ 
06 PS & 520 \\ 
07 Other & 40 \\
\hline
\end{tabular}
\end{center}
\label{table:images categ}
\end{table}

\section{Proposed Approach}
\begin{center}
\begin{figure*}[h]
\centerline{\includegraphics[width=120mm,height=25mm]{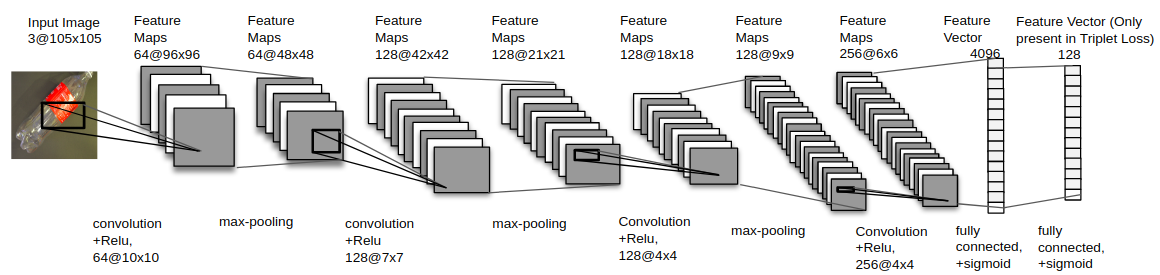}}
\caption{Convolutional Neural Network Structure.}
\label{cnn}
\end{figure*}
\end{center}

We learn image representations via a supervised metric-based approach with siamese and triplet loss neural networks, and then use the trained network weights for evaluation. For both the approaches, the CNN architecture used is inspired by \cite{Koch} and shown in Fig. \ref{cnn}. The siamese and triplet loss networks are explained in the next subsections. 
\vspace*{-10mm}
\subsection{Siamese Network}
\begin{figure}[h]
\centerline{\includegraphics[scale=0.5,width=68mm]{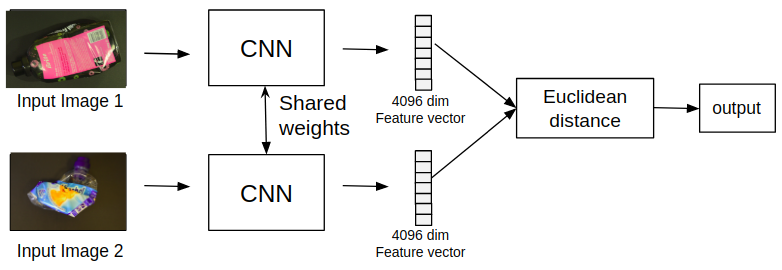}}
\caption{Siamese Network Structure}
\label{siamese}
\end{figure}

Two images are taken as the input and the label is set to 0 if they belong to the same class and 1 otherwise. Both images are passed through two identical CNN networks which have shared weights as shown in Fig. \ref{siamese}. The resultant output is a 4096 feature vector for each image. Then the Euclidean distance between the two feature vectors is computed, passed through a sigmoid layer and the resultant cross-entropy loss is propagated backward to both the identical networks.

\vspace*{-5mm}
\subsection{Triplet Loss} 
The architecture consists of three identical networks with shared weights. One image at random is selected and it is referred to as the anchor image. The other two images are called as positive (image of the same class) and negative (image of another class) images. All the three images are passed into the three identical CNN networks. The CNN network used is identical to the siamese, but an additional fully-connected layer of 128-dim is added after the last fully-connected layer of the siamese network. The loss is computed in equation (\ref{eq_triplet}) and it is propagated backward into each of the three identical networks. 
\begin{equation}
\\||f(x^a) - f(x^p)||^2_2 - ||f(x^a) - f(x^n)||^2_2 + 0.4 \label{eq_triplet}
\end{equation}

where $f(x^a)$, $f(x^p)$, $f(x^n)$ represent the anchor, positive and negative embedding respectively obtained after passing the corresponding images through the CNN. 
\begin{figure}[H]
\centerline{\includegraphics[width=68mm]{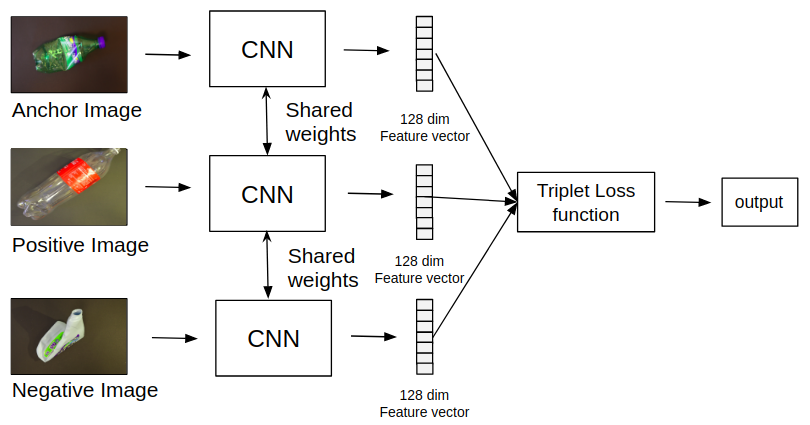}}
\caption{Triplet Network Structure}
\label{triplet}
\end{figure}

\section{Experiments}
\subsection{Training}
Siamese network is trained for 50 epochs, while the triplet loss network is trained for 100 epochs. In both the networks 5000 instances are generated per epoch. In both the cases \mbox{Stochastic Gradient Descent} optimizer is used with a learning rate of 0.001 and momentum 0.9. 
\subsection{Evaluation and Results}
Two methods for evaluation are used. The first one is \mbox{1-shot 5-way accuracy} in which 400 random sample images are evaluated. For each sample image, 5 test images are chosen randomly, one from each category. These 5 images are each paired with the sample image and sent as an input into the network. The sample image is assigned to the category of the test image for which the network's output is the minimum. Count of correctly classified images is incremented if the label of the sample image matches the assigned category.
In the second method, the features from the last fully-connected layer are extracted and stored in a database. Each image in the database is compared to all other images using euclidean distance, and accuracy is measured using the \mbox{K-nearest Neighbour algorithm.} Table \ref{one shot accuracy} and \mbox{Table \ref{knn accuracy}} show the accuracy obtained from \mbox{the 1-shot 5-way} and KNN \mbox{evaluation} methods respectively. 
\begin{table}[H]
\caption{1-Shot 5-way  Accuracy}
\begin{center}
\begin{tabular}{@{} c|c @{}}
\hline
\textbf{Method} & \textbf{Accuracy} \\ \hline
Siamese & 99.50\\
Triplet Loss & 99.0\\
\hline
\end{tabular}
\end{center}
\label{one shot accuracy}
\end{table}
\vspace*{-5mm}
\begin{table}[htbp]
\addtolength{\parskip}{-1mm}
\caption{KNN Accuracy}
\begin{center}
\begin{tabular}{@{} c|c|c|c|c|c|c|c @{}}
\hline
\textbf{Method} & \textbf{KNN} &\textbf{PET} & \textbf{PE-HD}
&\textbf{PP} & \textbf{PS} & \textbf{Other} & \textbf{Average} \\ \hline
Siamese & K = 3 & 99.59 & 99.66 & 99.35 & 100 & 100 & 99.72\\
Siamese & K = 5 & 99.4 & 99.83 & 99.00 & 100 & 100 & 99.64\\
Siamese & K = 7 & 99.59 & 99.66 & 99.20 & 100 & 100 & 99.69\\
TripletLoss & K = 3 & 99.68 & 100 & 99.30 & 100 & 100 & \textbf{99.79}\\
TripletLoss & K = 5 & 99.50 & 100 & 98.5 & 98.5 & 100 & 99.56\\
TripletLoss & K = 7 & 99.40 & 100 & 98.40 & 99.8 & 100 & 99.52\\
\hline
\end{tabular}
\end{center}
\label{knn accuracy}
\end{table}

\vspace{-5mm}

\section{Conclusion and Future Work}
We achieved a maximum accuracy of 99.79 on the WaDaBa database across all the 5 plastic categories without image augmentation. Further work will focus on object localization followed by classification in case of multiple objects within a single image.

\end{document}